\documentclass{article}
\usepackage{spconf,amsmath,graphicx}
\usepackage{stfloats}
\usepackage{color}
\usepackage{subcaption}
\usepackage{amssymb}
\usepackage{bbding}
\usepackage{multirow,booktabs}
\usepackage{bm}
\usepackage{cite}

\title{Self-Supervised Spatially Variant PSF Estimation\\for Aberration-Aware Depth-from-Defocus}

\name{Zhuofeng Wu, Yusuke Monno, and Masatoshi Okutomi}
\address{Tokyo Institute of Technology, Tokyo, Japan }

\begin{document}
\ninept

\maketitle

\begin{abstract}
In this paper, we address the task of aberration-aware depth-from-defocus (DfD), which takes account of spatially variant point spread functions~(PSFs) of a real camera. To effectively obtain the spatially variant PSFs of a real camera without requiring any ground-truth PSFs, we propose a novel self-supervised learning method that leverages the pair of real sharp and blurred images, which can be easily captured by changing the aperture setting of the camera. In our PSF estimation, we assume rotationally symmetric PSFs and introduce the polar coordinate system to more accurately learn the PSF estimation network. We also handle the focus breathing phenomenon that occurs in real DfD situations.  Experimental results on synthetic and real data demonstrate the effectiveness of our method regarding both the PSF estimation and the depth estimation.

\end{abstract}

\begin{keywords}
Depth-from-defocus, point spread function estimation, self-supervised learning
\end{keywords}

\section{Introduction}
\label{sec:intro}
Depth-from-defocus (DfD) is one of the well-adopted depth sensing techniques and estimates the scene depth from focal stack images based on the relationship between the depth and the defocus amount. 
Existing DfD methods~\cite{hiura1998depth, suwajanakorn2015depth, subbarao1994depth, song2018depth, maximov2020focus, wang2021bridging, yang2022deep} often assume a simplistic spatially invariant point spread function~(PSF) due to the difficulty in acquiring spatially variant PSFs of a real camera with non-ideal optics including lens distortions and optical aberrations. However, the assumption of a spatially invariant PSF is detrimental to deep-learning-based DfD methods, since they are typically trained using synthetic focal stack images assuming accurate PSFs of a real camera.

Existing spatially variant PSF estimation methods are classified as optimization-based methods~\cite{mannan2016blur, mannan2016good, joshi2008psf, mosleh2015camera, delbracio2012non} and learning-based methods~\cite{yang2023aberration, tseng2021differentiable}. The optimization-based methods use a real blurred image of a known calibration pattern to derive the PSF of each pixel that minimizes the difference between the real blurred pattern and the synthetically blurred pattern by the PSF. The learning-based methods train a neural network to infer PSFs of arbitrary pixels assuming that ground-truth PSFs are available for some of the image pixels, which are typically obtained using detailed lens parameters. These existing methods still have some limitations in practicability, such as the necessity of precise pattern alignment and ground-truth PSFs. 

\begin{figure}[t]

  \centering
  \centerline{\includegraphics[width=8.5cm]{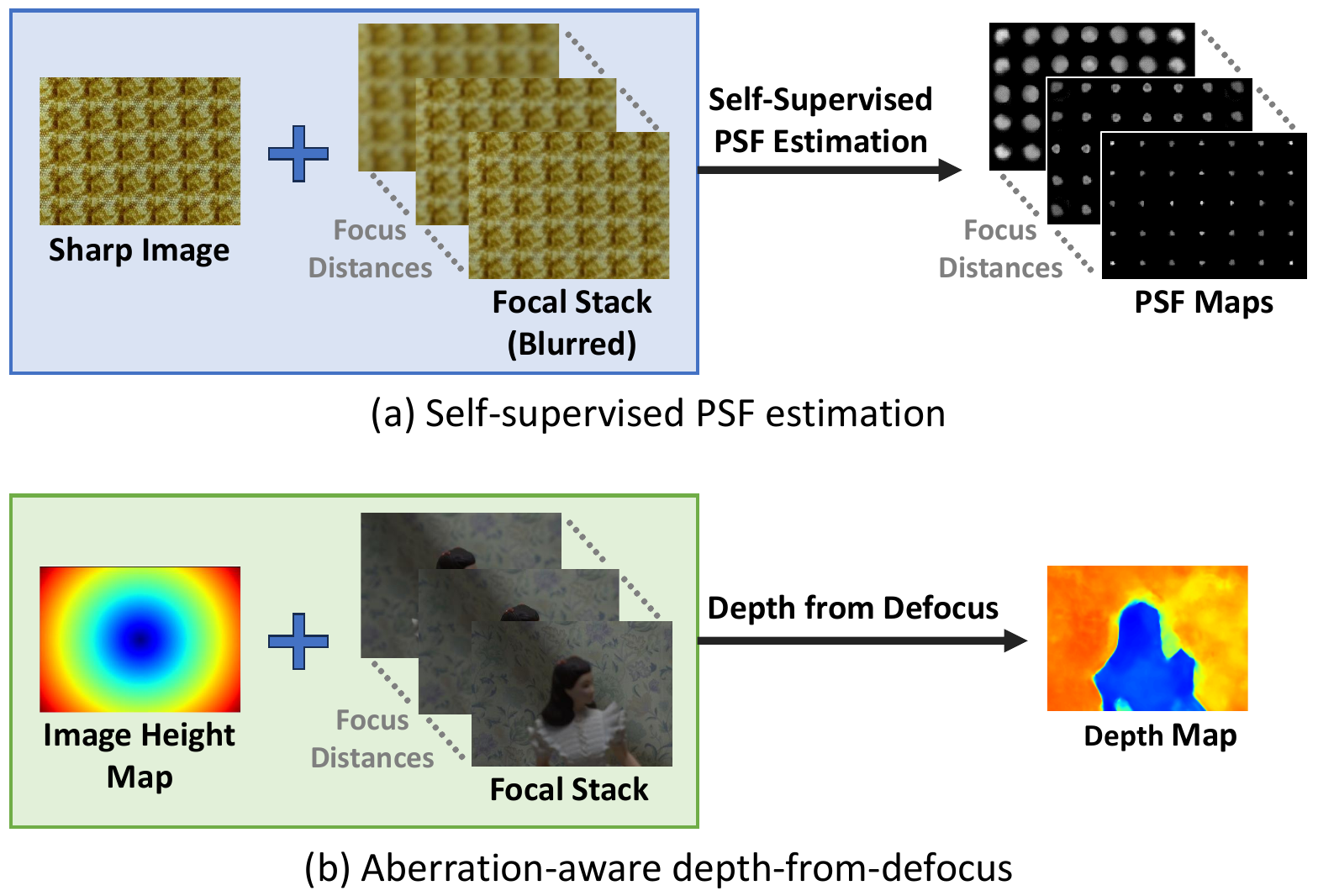}}
  \caption{The outline of this work. (a) We propose a novel self-supervised learning method for estimating spatially variant PSFs for each focus distance of the focal stack based on the pair of real sharp and blurred images. The obtained PSFs are used to generate synthetic focal stack images for depth-from-defocus~(DfD) network training. (b) We train a DfD network with additional image height map information to make the network learn the depth map considering the spatial variance of the PSFs.}
\label{fig:pipeline}

\end{figure}

In this paper, we propose a novel deep-learning-based spatially variant PSF estimation method. For training a PSF estimation network~(PSF-Net), we leverage the pair of real sharp and real blurred images, as shown in Fig.~\ref{fig:pipeline}(a). This brings us two main benefits: (i)~Our method is self-supervised without requiring ground-truth PSFs because the pair of sharp and blurred images can be conveniently captured using a real camera by adjusting the aperture size. (ii)~Our method is image alignment-free and naturally handles the effect of focus breathing, which is the phenomenon that the scale of image magnification (i.e., field-of-view) slightly differs according to the focus distance (details are in Sec.~\ref{sec:psfnet}).
In our PSF-Net training, we introduce the polar coordinate PSF model~\cite{maeda2005integrating} to incorporate the rotation symmetry of the PSFs, which improves the accuracy of the PSF estimation. Using the estimated spatially variant PSF maps for each focus distance of the focal stack, we generate synthetic focal stack images to train a DfD network, where we additionally provide the image height~(IH) information as a positional input to make the DfD network learn the depth considering the spatial variations of the PSFs, as illustrated in Fig.~\ref{fig:pipeline}(b). We experimentally demonstrate the effectiveness of our method in terms of both the PSF estimation and the depth estimation.

\section{Proposed Method}
\label{sec:method}
\subsection{Training Data Capturing for PSF Estimation}
For collecting the training data for PSF estimation, we capture the sharp and blurred images of a front-parallel plane with an arbitrary texture, which can be realized by a high-resolution display monitor. The initial focus distance of the camera is represented as ${f_d}_1$ and the camera F-value is set to its minimum value $F_1$ to realize a large depth-of-field. Then, the sharp image $I({f_d}_1, F_1, d)$ can be captured, where $d$ is the distance from the camera to the plane, i.e. the scene depth.
Then, the camera's F-value is changed to $F_2$ used for the subsequent DfD task. By using the focus bracketing function~\cite{ray2002applied}, the focal stack consisting of a series of blurred images with increasing focus distances (${f_d}_1, \cdots, {f_d}_N$) can be captured. The sharp image $I({f_d}_1, F_1, d)$ is then paired with every blurred image $\{I({f_d}_1, F_2, d), \cdots, I({f_d}_N, F_2, d)\}$ to construct a data pair for each focus distance. This process is repeated while varying the plane texture and depth to form the dataset.

\subsection{Polar Coordinate PSF Model}
\begin{figure}[tb]

  \centering
  \centerline{\includegraphics[width=8.5cm]{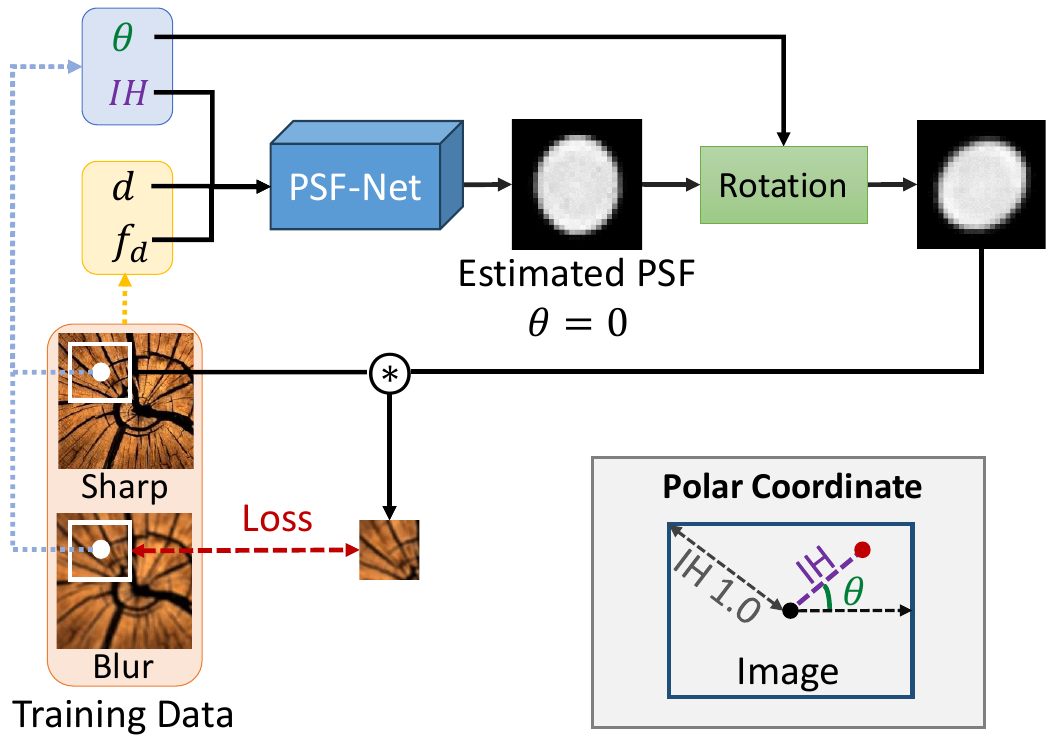}}
\caption{The overall flow of our self-supervised PSF-Net training.}
\label{fig:psfnet}
\end{figure}

Although spatially variant PSFs are naively defined for each pixel ($x,y$), they require a large amount of training data sufficient for learning the PSF of each pixel. Thus, to more efficiently represent the PSFs, we assume rotational symmetry of the lens, whereby the PSF's shape varies only along the image height~(IH), which is defined as the radius distance from the image center~\cite{tseng2021differentiable,maeda2005integrating}, as shown in the lower right part of Fig.~\ref{fig:psfnet}. The maximum height (i.e., $IH=1.0$) is defined as the distance from the image center to the image corner. The PSF's orientation is determined by the polar angle $\theta$ and increases in a counterclockwise direction from the polar axis. 

\subsection{Self-Supervised PSF Estimation Network Training}
\label{sec:psfnet}

Figure~\ref{fig:psfnet} illustrates the overall flow of our self-supervised PSF-Net training. Given a training pair of sharp and blurred images $\{I({f_d}_1, F_1, d), I({f_{d_n}}, F_2, d)\}$ sampled from the training dataset, a random patch centered at the pixel ($IH, \theta$) is used for the supervision. According to this, the PSF-Net inputs the image height $IH$, the scene depth $d$, and the camera’s focus distance $f_d$, and estimates the corresponding PSF. Since we assume rotationally symmetric PSFs, the orientation of the output PSF is assumed to be zero (i.e., $\theta=0$). Then, the output PSF is rotated by $\theta$ and then convolved with the sharp image patch centered at the pixel ($IH, \theta$) to generate the synthetically blurred patch. The loss is then calculated as the difference between the synthetic blurred patch and the real blurred patch captured for corresponding $f_{d}$ and $d$. These processes are performed while changing the training pair sample at each iteration. Since the set of focus distances used for the subsequent DfD task can be pre-determined according to the camera setup, we train PSF-Net for each focus distance separately.

\begin{figure}[t!]

  \centering
  \centerline{\includegraphics[width=8.5cm]{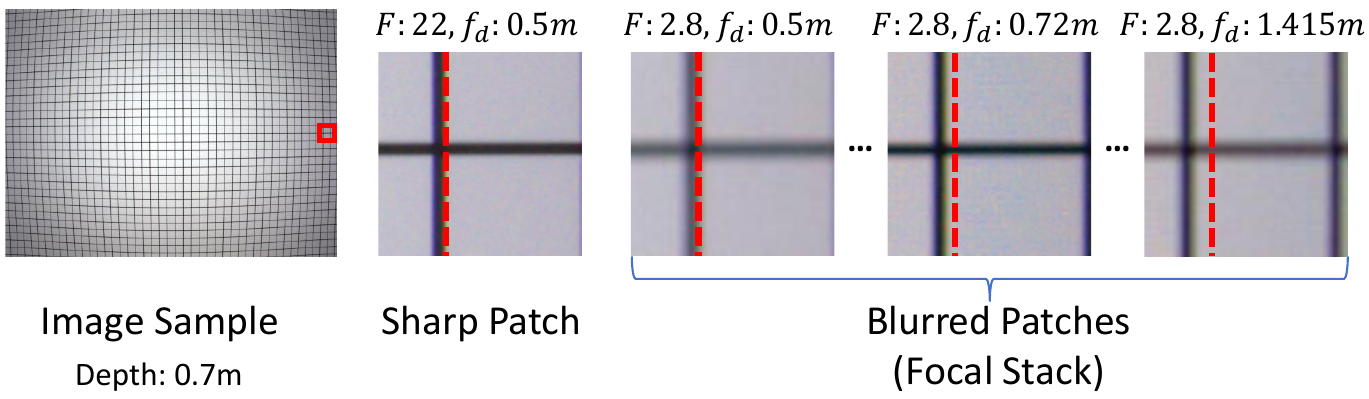}}

\caption{Examples of the image misalignments across the real focal stack images caused by the focus breathing phenomenon.}
\label{fig:breath}
\end{figure}

As shown in Fig.~\ref{fig:breath}, the image misalignments are observed along the focus distances of the real focal stack. This phenomenon, known as focus breathing, arises due to slight alterations of the field of view as the focus distance changes~\cite{rowlands2017physics}.
In our training procedure, the PSFs of each focus distance in (${f_d}_1, \cdots, {f_d}_N$)  are trained according to a single sharp image corresponding to the first focus distance ${f_d}_1$, meaning that the image misalignment by the focus breathing is naturally handled by the different shapes and gravity centers of the estimated PSFs for different focus distances. 

For PSF-Net, we employ a standard UNet architecture~\cite{ronneberger2015u}.

The loss function consists of three terms as 
\begin{equation}
\mathcal{L} = \mathcal{L}_{recon} + \alpha \mathcal{L}_{smooth} + \beta \mathcal{L}_{radial}
\label{formula: 1}
\end{equation}
where $\mathcal{L}_{recon}$ is a reconstruction loss, $\mathcal{L}_{smooth}$ is a smoothness loss, $\mathcal{L}_{radial}$ is a radial gradient loss, and $\alpha$ and $\beta$ are balancing weights, which are experimentally set to 1 and 10, respectively. The reconstruction loss is defined as
\begin{equation}
\mathcal{L}_{recon} = L_{1}(\bm{b}, \bm{s} \circledast \bm{p}(\theta))
\label{formula: 1}
\end{equation}
where $\bm{b}$ is the patch of the real blurred image, $\bm{s}$ is the corresponding patch in the sharp image, $\bm{p}(\theta)$ is the estimated PSF kernel after the rotation by $\theta$, and $\circledast$ represents the 2D convolution. 

To mitigate the impact of potential noise and derive  smooth PSFs, a smoothness regularization is applied as
\begin{equation}
\mathcal{L}_{smooth} = \frac{1}{N_p} \sum^{N}_{i}|\delta_{x}p_{i}|+|\delta_{y}p_{i}|
\label{formula: 1}
\end{equation}
where $\delta_{x}$ and $\delta_{y}$ are horizontal and vertical gradient calculation operations, $p_i$ is $i$-th pixel in the PFS kernel, and $N_p$ is the total number of the pixels in the PSF kernel. 

In addition, we introduce a radial gradient loss to encourage the network to estimate the PSF that gradually decreases from the center to the surroundings, thus preventing the PSF from being a donut-shaped form. The loss is defined as
\begin{equation}
\mathcal{L}_{radial} = \frac{1}{K} \sum^{K-1}_{j=0}T(p_{\Omega_{j+1}}-p_{\Omega_{j}}),
\label{formula: 1}
\end{equation}
where $p_{\Omega_{j}}$ denotes the average value within the center $(2j+1) \times (2j+1)$ patch of the PSF and $K$ represents the PSF kernel radius, which is set to 12 in our experiments. $T$ is a piecewise function, which is employed to pose a cost only if the average value becomes larger along the radial direction as follows.

\begin{equation}
T(\lambda)=\left\{
	\begin{aligned}
	\lambda \quad \lambda \geq 0\\
	0 \quad \lambda<0\\
	\end{aligned}
	\right
	.
\label{formula: 1}
\end{equation}

\subsection{Depth-from-Defocus Network Training}
\label{sec:dfdnet}
Using the trained PSF-Net for each focus distance $f_d$, we can generate the PSF for an arbitrary ($IH, \theta$), i.e., pixel position,  and scene depth $d$. Thus, we can synthesize input focal stack images from a ground-truth depth map by using an existing synthetic data generation method~\cite{wu2022realistic, lu2021self, lee2019deep}. Using the generated synthetic dataset, we next train a representative DfD network, such as DualDfDNet~\cite{song2018depth} and DefocusNet~\cite{maximov2020focus}. To consider the spatial variations of the PSFs in a DfD network, we provide the IH map, where each pixel has a corresponding IH value (see Fig.~\ref{fig:pipeline}(b) for illustration), as additional information. Specifically, for DualDfDNet, the IH map is concatenated with the features derived from the feature extractor, which comprises three weight-shared convolutional layers. For DefocusNet, we simply concatenate the IH map with input RGB channels.

\begin{table}[t!]
\small
 \caption{Quantitative evaluations on synthetic data.}
 \label{table:syn-result}
 \centering
 \setlength{\tabcolsep}{0.8mm}{
  \begin{tabular}{c|ccc|c} 
  \hline
  Methods & Supervised & \begin{tabular}[c]{@{}c@{}}Self-\\Supervised\end{tabular} & \begin{tabular}[c]{@{}c@{}}Spatially\\Variant\end{tabular}  & \begin{tabular}[c]{@{}c@{}}MAE\\($\times 10^{-3}$)\end{tabular}\\
  \hline
  Gaussian~\cite{subbarao1988depth} & \checkmark & &  & 1.99\\
  Tseng~\cite{tseng2021differentiable}& \checkmark &  & \checkmark & 1.26\\
  Yang\cite{yang2023aberration}&  \checkmark & &  \checkmark & 1.20\\
  Ours &  &\CheckmarkBold & \checkmark & 1.28\\
  \hline
  \end{tabular}}
\end{table}

\begin{figure}[t!]

  \centering
  \centerline{\includegraphics[width=8.5cm]{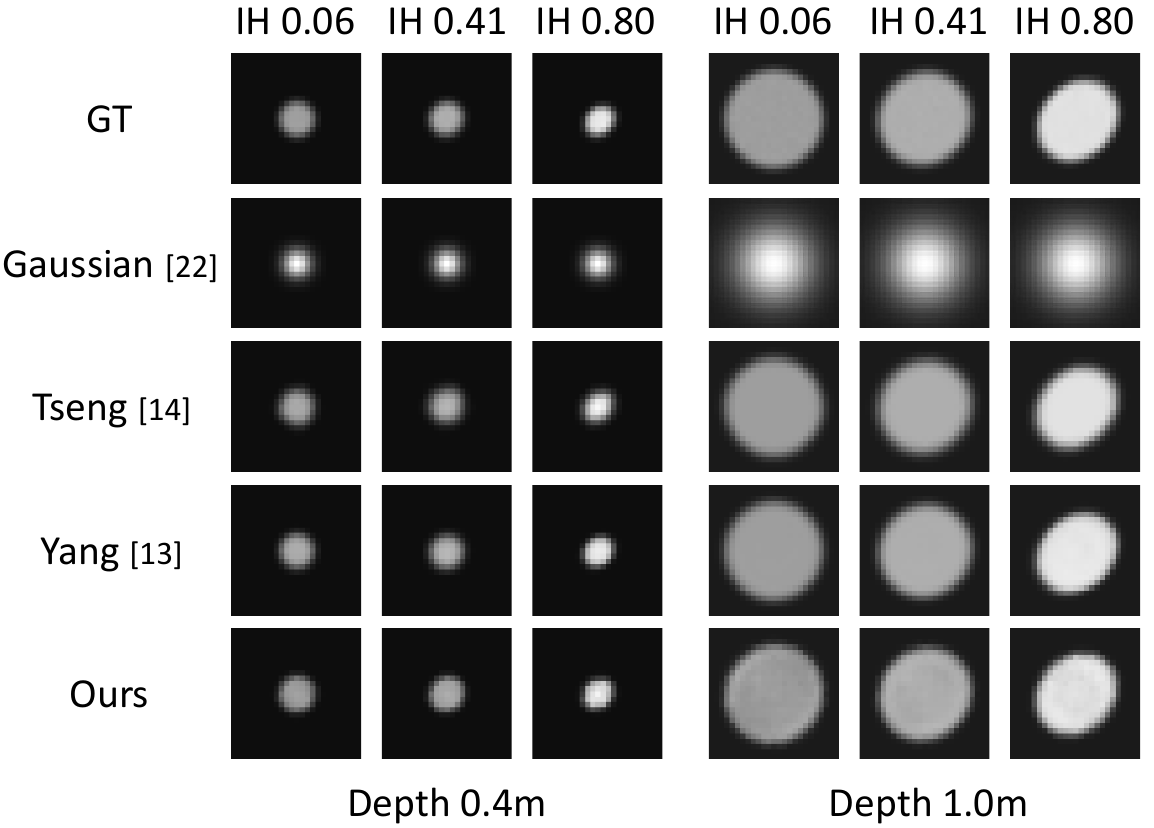}}

\caption{Visual comparison of the PSF estimation results ($\theta=45^\circ$) on synthetic data.}
\label{fig:syn-result}

\end{figure}

\section{Experimental Results}
\label{sec:experiments}
\subsection{PSF Estimation Results on Synthetic Data }

We first evaluate the PSF estimation results using ideal synthetic data. The synthetic data for our PSF-Net training were generated by the 3D graphics software, Blender~\cite{hess2013blender}. 
We set the camera's focal length to 20mm. The F-value was set to six for the blurred image (i.e., $F_2=6$). The sharp image was ideally obtained without any blurring (i.e., $F_1=\infty$). 
In the simulation, we evaluated the PSF-Net trained for the focus distance of 0.3 meters. 
During the image rendering for the sharp and blurred image pair, a front-parallel textured 3D plane, where the plane texture was randomly selected from the DTD texture dataset~\cite{cimpoi2014describing}, was captured with $501 \times 501$ pixel resolution while randomly changing the plane depth at 180 positions, resulting in a total of 180 sharp and blurred image pairs. Using those 180 image pairs as training data, we trained PSF-Net to estimate the PSFs for arbitrary pixel positions and depths. 

We compared our method with three existing methods: Gaussian PSF fitting \cite{subbarao1988depth}, Tseng's method \cite{tseng2021differentiable}, and Yang's method~\cite{yang2023aberration}, whose properties are summarized in Table~\ref{table:syn-result}.
Since all of the compared methods are supervised methods using ground-truth PSFs, we obtained them using Blender for sampled $17\times17$ pixels for each depth, which are uniformly distributed among the whole $501\times501$ pixels. Then, we randomly selected 20 depths for the supervision, resulting in a total of $17\times17\times20=5780$ ground-truth PSFs. Using those PSFs as training data, the spatially invariant Gaussian PSF fitting~\cite{subbarao1988depth} and the spatially variant PSF training~\cite{yang2023aberration,tseng2021differentiable} were performed. Then, the PSF estimation results of all the methods, including ours, were evaluated on testing data, which are ground-truth PSFs for sampled $16\times16$ pixels at eight depths, i.e., a total of 2,048 PSFs, not included in the training data. 

Table~\ref{table:syn-result} presents the mean absolute errors (MAEs) of the PSF estimation results and Fig.~\ref{fig:syn-result} shows the visualization of the PSF results ($\theta=45^\circ$) at the depths of 0.4 and 1.0 meters. The spatially invariant Gaussian PSF significantly deviates from the ground-truth PSF, resulting in a larger numerical error compared with the spatially variant methods. From both the numerical and the visual results, we can confirm that our method demonstrates the ability to infer PSFs that closely resemble the ground-truth PSFs and achieves comparable performance to the other supervised methods, even without relying on any ground-truth PSFs for the training.

\begin{table}[t!]
\small
 \caption{Results of ablation study.}
 \label{table:ablation}
 \centering
 \setlength{\tabcolsep}{1.0mm}{
  \begin{tabular}{cc|c} 
  \hline
   Spatially Variant & Positional Input & MAE ($\times 10^{-3}$)\\
  \hline
     & none & 1.89\\
  \checkmark & (x, y) & 1.44\\
  \checkmark & IH & 1.28\\
  \hline
  \end{tabular}}
\end{table}

\label{sec:ablation}
Table~\ref{table:ablation} shows the results of an ablation study. To confirm the effectiveness of our spatially variant PSF estimation with the IH input (the third row), we first compared a spatially invariant version of our method (the first row). In this version, we removed all positional inputs and only provided depth and focus distance as input variables to estimate a spatially invariant PSF. 
We also compared the version in which we replaced the positional input of IH with a standard (x, y) pixel coordinate and estimated PSFs for every pixel without assuming the rotational symmetry of PSFs (the second row). The results in Table~\ref{table:ablation} show that both the spatially variant versions outperform the spatially invariant version. The comparison of the second and the third rows provides substantial support for the effectiveness of the polar coordinate PSF model assuming the rotational symmetry. This enhanced performance can be attributed to the reduction in input and output data variability, consequently leading to increased robustness and accuracy during the training process.

\subsection{PSF Estimation Results on Real Data}
\label{sec:realpsf}
We applied our PSF estimation method to an Olympus OM-D E-M5 Mark III camera with M.ZUIKO DIGITAL ED 12-40mm F2.8 PRO lens, where we fixed the focal length to 12mm. The F-values were set as 22 and 2.8 for capturing sharp and blurred images, respectively. The focus bracketing function was utilized to capture focal stack images at the focus distances of [0.5, 0.59, 0.72, 0.94, 1.415] meters. We set the target depth range of our PSF estimation and subsequent depth estimation from the minimum to the maximum of the focus distances (i.e., [0.5, 1.415] meters). To collect the training sharp and blurred image pairs for PSF estimation, we used a 4K-65-inch display monitor and captured front-parallel scenes since estimating the PSFs assuming arbitrarily spatially varying depths is infeasible in practice. We positioned the camera at 20 depth positions from the monitor, capturing three displayed textures at each depth. The textures were randomly selected from the DTD dataset~\cite{cimpoi2014describing}.

\begin{figure}[t!]

  \centering
  \centerline{\includegraphics[width=8.5cm]{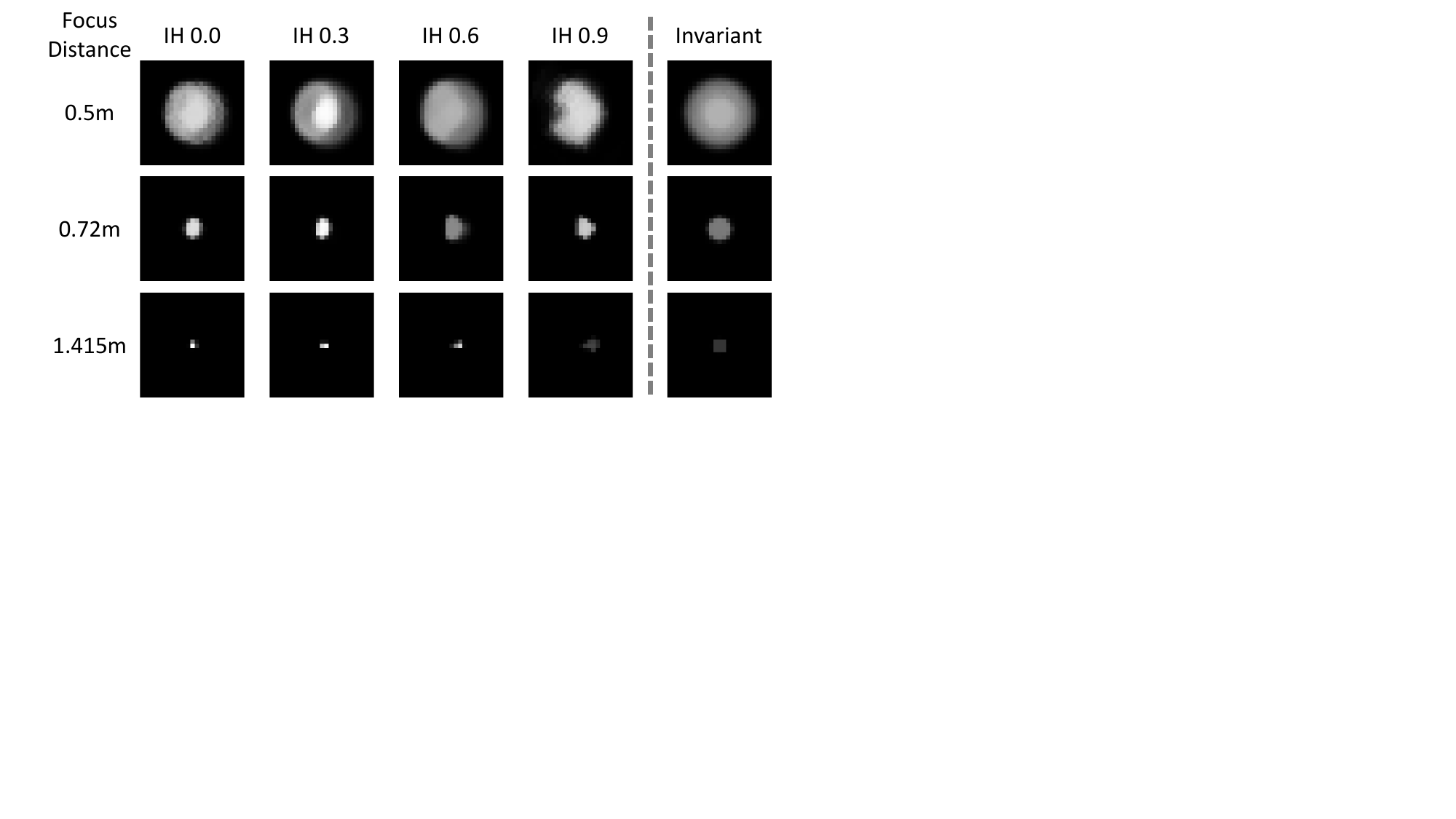}}
 \vspace{-1mm}
\caption{PSF estimation results ($\theta=0^\circ$) using a real Olympus camera for the depth of 1.0 meters.}
\label{fig:realpsf}
\end{figure}

In Fig.~\ref{fig:realpsf}, we show the estimated PSFs corresponding to the three focus distances at the depth of 1.0 meters. It can be observed that, as the IH increases, the PSFs of the same focus distance deviate further from a circular shape, suggesting that the PSFs of a real camera are highly spatially variant and change their shapes from the image center to the image surroundings according to optical aberrations and lens distortions. Thus, a single spatially invariant PSF estimated using the same training data (the last column) cannot sufficiently represent the PSFs of a real camera.
Also, as discussed in Sec.~\ref{sec:psfnet}, our method can handle the focus breathing phenomenon. This can be observed in the PSFs of different focus distances for the same IH, where the gravity centers of the PSFs are gradually shifted according to the focus distance. This is especially apparent for a large IH, such as $IH=0.9$, because the focus breathing causes the image misalignment much more at the image surroundings. 

\subsection{Depth-from-Defocus Results on Real Data}
We applied the same Olympus camera and settings as the PSF estimation to the DfD task, where we input focal stack images of five focus distances. To generate training data for DfD, we employed the 2-plane data generation method in~\cite{wu2022realistic} and generated 5,000 pairs of the input focal stack and the ground-truth depth map. Then, we trained two DfD networks, DualDfDNet~\cite{song2018depth} and DefocusNet~\cite{maximov2020focus}, using the generated training dataset.

To numerically evaluate the estimated depth by using the real Olympus camera, we captured a testing dataset of simple front-parallel one-plane textured scenes using the display monitor, as in Sec.~\ref{sec:realpsf}. We obtained the pair of input focal stack and ground-truth monitor depth at 20 different depth settings, each with three distinct textures, resulting in a total of 60 one-plane scenes. For these testing one-plane scenes, we evaluated three versions of the method, as shown in Table~\ref{table:real1p}. 

The first version uses the original DualDfD/DefocusNet network architecture and the training dataset generated by using the spatially invariant PSF estimated as in the last column of Fig.~\ref{fig:realpsf}. The second version uses the original DualDfD/DefocusNet architecture and the training dataset generated by using the spatially variant PSFs estimated by our method. The third version is our final proposal, where the IH map (see Fig.~\ref{fig:pipeline}(b) for illustration) is added to the input of DualDfD/DefocusNet, as explained in Sec.~\ref{sec:dfdnet}. 
Table~\ref{table:real1p} shows that training the DfD networks with the dataset using the spatially invariant PSF leads to poor performance. A notable improvement can be observed when the DfD networks are trained with the dataset using our spatially variant PSFs. The use of the IH map as the network input further improves the performance. From those results, we can confirm the effectiveness of the spatially variant PSFs and the IH map input to the real DfD problem.

\begin{figure}[t!]

  \centering
  \centerline{\includegraphics[width=8.5cm]{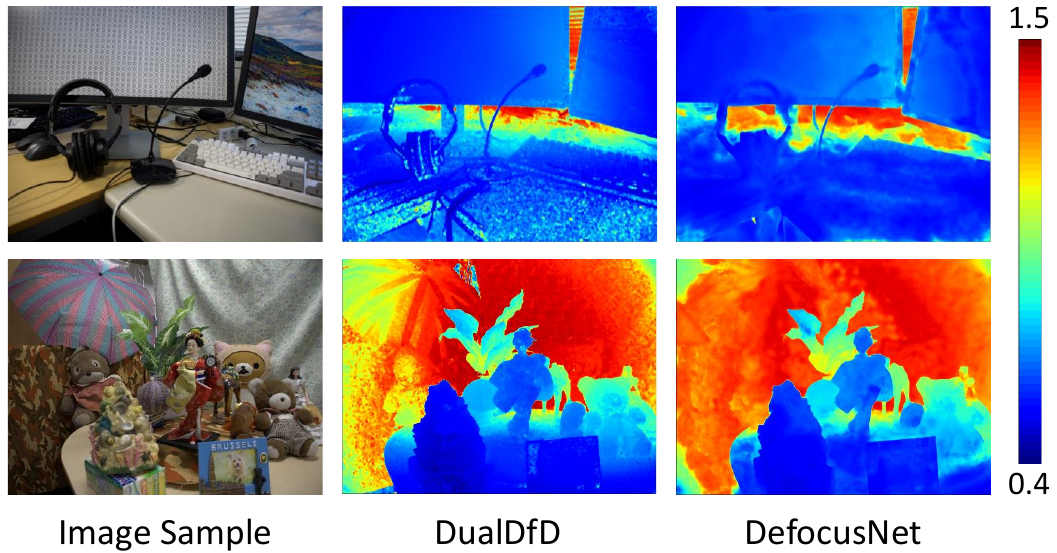}}
 \vspace{-1mm}
\caption{Depth estimation results on two real scenes.}
\label{fig:realscene}
\end{figure}

The examples of the depth estimation results for two real scenes are shown in Fig.~\ref{fig:realscene}. Both DfD networks trained with our approach can reasonably well estimate the depth maps of the complex real scenes. Comparing the results of the two networks, DefocusNet tends to generate smoother and overall better depth maps, while DualDfD tends to better estimate depth boundaries, such as microphones in the scene on the top.

\begin{table}[t!]
\small
 \caption{Quantitative evaluations on real one-plane data.}
 \label{table:real1p}
 \centering
 \setlength{\tabcolsep}{1.0mm}{
  \begin{tabular}{c|cc|c} 
  \hline
  Methods &  Spatially Variant & IH Input & MAE \\
  \hline
  \multirow{3}*{DualDfD}    &            &  & 0.2351\\
                            & \checkmark &  & 0.1000\\
                            & \checkmark & \checkmark & 0.0503\\
  \hline
  \multirow{3}*{DefocusNet} &            &  & 0.0997\\
                            & \checkmark &  & 0.0576\\
                            & \checkmark & \checkmark & 0.0454\\
  \hline
  \end{tabular}}
\end{table}

\section{Conclusion}
\label{sec:conclusion}
In this paper, we have proposed a novel self-supervised PSF estimation method that eliminates the requirements of ground-truth PSFs, which are hard to obtain for a real camera with non-ideal optics. Specifically, we have utilized the pair of sharp and blurred images directly captured by a real camera, which enables us to estimate the spatially variant PSFs for all pixel locations via self-supervision. We have also introduced the polar coordinate PSF model to represent rotationally symmetric PSFs and provided the IH map to a DfD network so that the network estimates the depth considering the spatial variance of the PSFs.
Experimental results on an ideal synthetic dataset have demonstrated that our method achieves comparable PSF estimation performance to existing supervised methods, even without any ground-truth PSFs. Experimental results on real data have demonstrated the effectiveness of our method in both the PSF estimation and the depth estimation. 
Our future work includes applying our method to a lens exhibiting more significant spatial PSF variations, such as fisheye lenses or microscopes.

\vfill\pagebreak

\bibliographystyle{IEEEbib}
\bibliography{main}

\end{document}